\documentclass[runningheads]{llncs}

\usepackage{graphicx}
\usepackage{tabularx}
\usepackage{color}
\usepackage{soul}
\usepackage{epstopdf}
\usepackage[latin1]{inputenc}

\usepackage{hyperref}
\usepackage{xstring}
\usepackage[square,numbers,sectionbib]{natbib}
\usepackage{cleveref}

\begin{document}
\title{FinEst BERT and CroSloEngual BERT: \\less is more in multilingual models}
\author{Matej Ul\v{c}ar \and Marko Robnik-\v{S}ikonja}
\institute{University of Ljubljana, Faculty of Computer and Information Science \\
Ve\v{c}na pot 113, Ljubljana, Slovenia \\
\{matej.ulcar, marko.robnik\}@fri.uni-lj.si}
%
%
%
%

\maketitle
\begin{abstract}
Large pretrained masked language models have become state-of-the-art solutions for many NLP problems. The research has been mostly focused on English language, though. While massively multilingual models exist, studies have shown that monolingual models produce much better results. We train two trilingual BERT-like models, one for Finnish, Estonian, and English, the other for Croatian, Slovenian, and English. We evaluate their performance on several downstream tasks, NER, POS-tagging, and dependency parsing, using the multilingual BERT and XLM-R as baselines. The newly created FinEst BERT and CroSloEngual BERT improve the results on all tasks in most monolingual and cross-lingual situations.
\end{abstract}

\section{Introduction}

In natural language processing (NLP), a lot of research focuses on numeric word representations. Static pretrained word embeddings like word2vec \cite{mikolov2013exploiting} 
are recently replaced by dynamic, contextual embeddings, such as ELMo \cite{Peters2018} and BERT \cite{Devlin2018}. These generate a word vector based on the context the word appears in, mostly using the sentence as the context.

Large pretrained masked language models like BERT \cite{Devlin2018} and its derivatives 
achieve state-of-the-art performance when fine-tuned for specific NLP tasks. The research into these models has been mostly limited to English and a few other well-resourced languages, such as Chinese Mandarin, French, German, and Spanish. However, two massively multilingual masked language models have been released: a multilingual BERT (mBERT) \cite{Devlin2018}, trained on 104 languages, and newer even larger XLM-RoBERTa (XLM-R) \cite{conneau2019unsupervised}, trained on 100 languages. While both, mBERT and XLM-R, achieve good results, it has been shown that monolingual models significantly outperform multilingual models \cite{virtanen2019multilingual,martin2019camembert}.
In our work, we reduced the number of languages in multilingual models to three, two similar less-resourced languages from the same language family, and English. The main reasons for this choice are to  better represent each language, and keep sensible sub-word vocabulary, as shown by \citet{virtanen2019multilingual}. We decided against production of monolingual models, because we are interested in using the models in multilingual sense and for cross-lingual knowledge transfer. By including English in each of the two models, we expect to better transfer existing prediction models from English to involved less-resourced languages.
Additional reason against purely monolingual models for less-resourced languages is the size of training corpora, i.e. BERT-like models use transformer architecture 
which is known to be data hungry.

We thus trained two multilingual BERT models: 
FinEst BERT was trained on Finnish, Estonian, and English, while CroSloEngual BERT was trained on Croatian, Slovenian, and English. In the paper, we present the creation and evaluation of these models, which required considerable computational resources, unavailable to most NLP researchers. 
We make the models which are valuable resources for the involved less-resourced languages publicly available\footnote{CroSloEngual BERT: \url{http://hdl.handle.net/11356/1317}\newline FinEst BERT: \url{http://urn.fi/urn:nbn:fi:lb-2020061201}}. 


\section{Training data and preprocessing}
\label{sec:datasets}
BERT models require large quantities of monolingual data. In \Cref{sec:corpora} we first describe the corpora used, followed by a short description of their preprocessing in \Cref{sec:preprocessing}.
\subsection{Datasets}
\label{sec:corpora}
We trained two new BERT models from five languages: Finnish, Estonian, Slovenian, Croatian and English. 
To obtain high-quality models, we used large monolingual corpora for each language, some of them unavailable to the general public. For English, large corpora are readily available and they are much larger than for other languages. However, high-quality English language models already exist and English is not the main focus of this research, we therefore did not use all available English corpora in order to prevent English from overwhelming the other languages in our models. Some corpora are available online under permissive licences, others are available only for research purposes or have limited availability. The corpora used in training are a mix of news articles and general web crawl, which we preprocessed and deduplicated.  
Details about the training set sizes are presented in \Cref{tab:trainingsize}, while their description can be found in works on the involved less-resourced languages, e,g.,  \cite{ular-robnikikonja:2020:LREC}.   

\begin{table}[tb]
\centering
\begin{minipage}[t]{0.48 \textwidth}
\begin{center}
\caption{The training corpora sizes in number of tokens and the ratios for each language. }
\label{tab:trainingsize}
\begin{tabular}{lrr}
Model & CroSloEngual & FinEst \\ \hline
Croatian & 31\% & 0\% \\
Slovenian & 23\% & 0\% \\
English & 47\% & 63\% \\
Estonian & 0\% & 13\% \\
Finnish & 0\% & 25\% \\
	\hline 
Tokens & $5.9\cdot10^9$ & $3.7\cdot10^9$ \\ \hline
\end{tabular} 
\end{center}
\end{minipage}
\hfill
\begin{minipage}[t]{0.48 \textwidth}
\begin{center}
\caption{The sizes of corpora subsets in millions of tokens used to create wordpiece vocabularies. }
\label{tab:wordpiece}
\begin{tabular}{lrr}
      && \\
      Language & FinEst & CroSloEngual \\
      \hline
      Croatian & / & 27  \\
      Slovenian & / & 28  \\
      English & 157 & 23  \\
      Estonian  & 75 & /  \\
      Finnish  & 97 & /  \\
      \hline
\end{tabular}
 \end{center}
\end{minipage}
\end{table}

\subsection{Preprocessing}
\label{sec:preprocessing}
Before using the corpora, we deduplicated them for each language separately, using the Onion (ONe Instance ONly) tool\footnote{\url{http://corpus.tools/wiki/Onion}}. We applied the tool on sentence level for those corpora that did have sentences shuffled, and on paragraph level for the rest. As parameters, we used 9-grams with duplicate content threshold of 0.9.

BERT models are trained on subword (wordpiece) tokens. We created a wordpiece vocabulary using bert-vocab-builder tool\footnote{\url{https://github.com/kwonmha/bert-vocab-builder}}, which is built upon tensor2tensor library \cite{tensor2tensor}. We did not process the whole corpora in creating the wordpiece vocabulary, but only a smaller subset. To balance the language representation in vocabulary, we used samples from each language. The sizes of corpora subsets are shown in Table \ref{tab:wordpiece}. The created wordpiece vocabularies contain 74,986 tokens for FinEst and 49,601 tokens for CroSloEngual model.



\section{Architecture and training}
\label{sec:training}
We trained two BERT multilingual models. FinEst BERT was trained on Finnish, Estonian, and English corpora, with altogether $3.7$ billion tokens.  CroSloEngual BERT was trained on Croatian, Slovenian, and English corpora with together $5.9$ billion tokens. 

Both models use bert-base architecture \cite{Devlin2018}, which is a 12-layer bidirectional transformer encoder with the hidden layer size of 768 and altogether 110 million parameters. We used the whole word masking for the masked language model training task. Both models are cased, i.e. the case information was preserved. We followed the hyper-parameters settings of \citet{Devlin2018}, except for the batch size and total number of steps. We trained the models for approximately 40 epochs with maximum sequence length of 128 tokens, followed by approximately 4 epochs with maximum sequence length of 512 tokens. The exact number of steps was calculated using the expression:
\[
s = \frac{N_{tok}\cdot E}{b\cdot \lambda}
\],
where $s$ is the number of steps the models were trained for, $N_{tok}$ is the number of tokens in the train corpora, $E$ is the desired number of epochs (in our case 40 and 4), $b$ is the batch size, and $\lambda$ is the maximum sequence length.

We trained FinEst BERT on a single Google Cloud TPU v3 for a total of 1.24 million steps where the first 1.13 million steps used the batch size of 1024 and sequence length 128, and the last 113 thousand steps used the batch size 256 and sequence length 512. 
Similarly, CroSloEngual BERT was trained on a single Google Cloud TPU v2 for a total of 3.96 million steps, where the first 3.6 million steps used the  batch size of 512 and sequence length 128, and the last 360 thousand steps were trained with the batch size 128 and sequence length 512. Training took approximately 2 weeks for FinEst BERT and approximately 3 weeks for CroSloEngual BERT.

\section{Evaluation}
\label{sec:evaluation}
We evaluated the two new BERT models on three downstream evaluation tasks available for the four involved less-resourced languages: named entity recognition (NER), part-of-speech tagging (POS), and dependency parsing (DP). We compared both models with BERT-base-multilingual-cased model (mBERT) on sensible languages, i.e. FinEst BERT was compared with mBERT on Finnish, Estonian, and English, while CroSloEngual BERT was compared with mBERT on Croatian, Slovenian, and English.

\subsection{Named Entity Recognition}
Named entity recognition (NER) task is a sequence labeling task, which tries to correctly identify and classify each token from an unstructured text into one of the predefined named entity (NE) classes or, if the token is not part of a NE, to classify it as not a named entity. Most common named entity classes are personal names, locations and organizations. We used various datasets, which do not cover the same set of classes. We therefore adapted the datasets to allow a more direct comparison between languages, by reducing them to the four labels they all have in common: PER (person), LOC (location), ORG (organization), and O (other). All tokens, which are not named entities or belong to any NE class other than person, location or organization, were labeled as 'O'.

For Croatian and Slovenian, we used data from hr500k \cite{LJUBEI16.340} and ssj500k \cite{ssj500k}, respectively. Not all sentences in ssj500k are annotated, so we excluded those that are not annotated. English dataset comes from CoNLL 2013 shared task \cite{tjongkimsang2003conll}. For Finnish we used Finnish News Corpus for NER \cite{ruokolainen2020finer}, and for Estonian dataset we used Nime\"{u}ksuste korpus \cite{estonian-ner}. The statistics of each dataset are shown in Table \ref{tab:nertags}.

\begin{table}[!h]
\begin{center}
\begin{tabular}{lrrrrr}
      & & & & \\
      Language & PER & LOC & ORG & Density & N \\ \hline
      Croatian & 10241 & 7445 & 11216 & 0.057 & 506457 \\ 
      English & 17050 & 12316 & 14613 & 0.146 & 301418 \\ 
      Estonian & 8490 & 6326 & 6149 & 0.096 & 217272 \\ 
      Finnish & 3402 & 2173 & 11258 & 0.087 & 193742 \\ 
      Slovenian & 4478 & 2460 & 2667 & 0.049 & 194667 \\ 
      \hline
\end{tabular}

 \caption{The number of tokens labeled with each label (PER, LOC, ORG), the density of these labels (their sum divided by the number of all tokens) and the number of all tokens (N) for datasets in all languages.}
\label{tab:nertags}
\end{center}
\end{table}

To evaluate the performance of BERT embeddings on the NER task we trained NER models using Huggingface's Transformer library, basing the code on their NER example\footnote{\url{https://github.com/huggingface/transformers/tree/master/examples/ner}}. We fine-tuned each of our BERT models with an added token classification head for 3 epochs on the NER data. We compared the results with BERT-base-multilingual-cased (mBERT) model, which we fine-tuned with exactly the same parameters on the same data.

\begin{table}[!h]
\begin{center}
\begin{tabular}{llrr}
      & & & \\
      Train lang & Test lang & mBERT & CroSloEngual \\
      \hline
      Croatian & Croatian & 0.795 & 0.894 \\
      Slovenian & Slovenian & 0.903 & 0.917 \\
      English & English & 0.940 & 0.949 \\ \hline
      Croatian & English & 0.793 & 0.866 \\
      English & Croatian & 0.638 & 0.798 \\
      Slovenian & English & 0.781 & 0.833 \\
      English & Slovenian & 0.736 & 0.843 \\
      Croatian & Slovenian & 0.825 & 0.908 \\
      Slovenian & Croatian & 0.755 & 0.847 \\
      \hline
\end{tabular}
 
 \caption{The results of NER evaluation task on Croatian, Slovenian, and English. The scores are average $F_1$ scores of the three named entity classes. A NER model was trained on "train language" dataset and tested on "test language" dataset using two different BERT models for all possible combinations of train and test languages.}
\label{tab:nercrosloeng}
\end{center}
\end{table}

We evaluated the models in a monolingual setting (training and testing on the same language) and a crosslingual setting (training on one language, testing on another). We present the results as macro average $F_1$ scores of the three NE classes, excluding 'O' label. Comparison between CroSloEngual BERT and mBERT is shown in Table~\ref{tab:nercrosloeng}, comparison between FinEst BERT and mBERT is shown in Table~\ref{tab:nerfinest}.

The difference in performance of each BERT on English data is negligible. In other languages, our models outperform the multilingual BERT, the difference is especially large in Croatian. In crosslingual setting, both FinEst BERT and CroSloEngual BERT show a significant improvement over mBERT, especially when one of the two languages is English. This leads us to believe that multilingual BERT models with fewer languages are more suitable for crosslingual knowledge transfer.

\begin{table}[!h]
\begin{center}
\begin{tabular}{llrr}
      & & & \\
      Train lang & Test lang & mBERT & FinEst \\
      \hline
      Finnish & Finnish & 0.922 & 0.959 \\
      Estonian & Estonian & 0.906 & 0.930 \\
      English & English & 0.940 & 0.942 \\ \hline
      Finnish & English & 0.692 & 0.810 \\
      English & Finnish & 0.770 & 0.901 \\
      Estonian & English & 0.765 & 0.815 \\
      English & Estonian & 0.762 & 0.839 \\
      Finnish & Estonian & 0.795 & 0.879 \\
      Estonian & Finnish & 0.839 & 0.912 \\   
      \hline
   
\end{tabular}
 
 \caption{The results of NER evaluation task on Finnish, Estonian, and English. The scores are average $F_1$ scores of the three named entity classes. A NER model was trained on "train language" dataset and tested on "test language" dataset using two different BERT models for all possible combinations of train and test languages.}
\label{tab:nerfinest}
\end{center}
\end{table}

\subsection{Part-of-speech tagging and dependency parsing}
We evaluated BERT models on two more classification tasks: part-of-speech (POS) tagging and dependency parsing. In the POS tagging task we attempt to correctly classify each token within a given set of grammatical categories (verb, adjective, punctuation, adverb, noun, etc.) Dependency parsing task attempts to predict the tree structure, representing the syntactic relations between words in a given sentence.

We trained classifiers on universal dependencies (UD) treebank datasets, using universal part-of-speech (UPOS) tag set. For Croatian, we used treebank by \citet{agic-ljubesic-2015-universal}. For English, we used A Gold Standard Dependency Corpus \cite{silveira14gold}. For Estonian, we used Estonian Dependency Treebank \cite{edt}, converted to UD. Finnish treebank used is based on the Turku Dependency Treebank \cite{fiTDT}, which was also converted to UD \cite{fiUD}. Slovenian treebank \cite{dobrovoljc17udtreebank} is based on the ssj500k corpus \cite{ssj500k}.

We used Udify tool \cite{kondratyuk-straka-2019-75} to train both POS tagger and dependency parsing classifiers at the same time. We finetuned each BERT model for 80 epochs on the treebank data. We kept the tool parameters at default values, except for "warmup\_steps" and "start\_step" values, which we changed to equal the number of training batches in one epoch.

We present the results of POS tagging as UPOS accuracy score in Table~\ref{tab:upos} and Table~\ref{tab:uposfinest}. The difference in performance between BERT models is very small on this task. FinEst and CroSloEngual BERTs perform slightly better than mBERT on all languages in monolingual setting, except Croatian, where mBERT and CroSloEngual BERT are equal. The differences are more pronounced in cross-lingual setting. When training on Slovenian, Finnish or Estonian data and testing on English data CroSloEngual and FinEst BERT significantly outperform mBERT. On the other hand, when training on English and testing Croatian, mBERT outperforms CroSloEngual BERT.

\begin{table}[!h]
\begin{center}
\begin{tabular}{llrr}
      && \\
      Train lang. & Test lang. & mBERT & CroSloEngual \\
      \hline
      Croatian & Croatian & 0.983 & 0.983  \\
      English & English & 0.969 & 0.972   \\
      Slovenian & Slovenian & 0.987 & 0.991  \\
      \hline
      English & Croatian   & 0.876 & 0.869  \\
      English & Slovenian  & 0.857 & 0.859  \\
      Croatian & English   & 0.750 & 0.756  \\
      Croatian & Slovenian & 0.917 & 0.934  \\
      Slovenian & English &  0.686 & 0.723  \\
      Slovenian & Croatian & 0.920 & 0.935  \\
      \hline
\end{tabular}
\caption{The embeddings quality measured on the UPOS tagging task, using UPOS accuracy score for FinEst BERT, CroSloEngual BERT and BERT-base-multilingual-cased (mBERT). }
\label{tab:upos}
 \end{center}
\end{table}

\begin{table}[!h]
\begin{center}
\begin{tabular}{llrr}
      && \\
      Train lang. & Test lang. & mBERT & FinEst \\
      \hline
      English & English & 0.969 & 0.970   \\
      Estonian & Estonian & 0.972 & 0.978  \\
      Finnish & Finnish & 0.970 & 0.981  \\
      \hline
      English & Estonian   & 0.852 & 0.878  \\
      English & Finnish  & 0.847 & 0.872  \\
      Estonian & English   & 0.688 & 0.808  \\
      Estonian & Finnish & 0.872 & 0.913  \\
      Finnish & English &  0.535 & 0.701  \\
      Finnish & Estonian & 0.888 & 0.919  \\
      \hline
\end{tabular}
\caption{The embeddings quality measured on the UPOS tagging task, using UPOS accuracy score for FinEst BERT, CroSloEngual BERT and BERT-base-multilingual-cased (mBERT). }
\label{tab:uposfinest}
 \end{center}
\end{table}

We present the results of dependency parsing task as unlabeled attachement score (UAS) and labeled attachment score (LAS). In monolingual setting CroSloEngual BERT shows improvement over mBERT on all three languages (Table \ref{tab:depparse1}) with the highest improvement on Slovenian and only a marginal improvement on English. FinEst BERT outperforms mBERT on Estonian and Finnish, with the biggest margin being on the Finnish data (Table \ref{tab:depparse2}). FinEst BERT and mBERT perform equally on English data.

In crosslingual setting, the results are similar to those seen on the POS tagging task. Major improvements of FinEst BERT and CroSloEngual BERT over mBERT in English-Estonian, English-Finnish and English-Slovenian pairs, minor improvements in Estonian-Finnish and Croatian-Slovenian pairs. Again, mBERT outperformed CroSloEngual BERT when dependency parser was trained on English data and tested on Croatian data.

\begin{table}[!h]
\begin{center}
\begin{tabular}{llrr|rr}
      Train & Test & \multicolumn{2}{c}{mBERT} & \multicolumn{2}{c}{CroSloEngual} \\
      language & language & UAS & LAS & UAS & LAS \\
      \hline
      Croatian & Croatian & 0.930 & 0.891 & 0.940 & 0.903  \\
      English & English & 0.917 & 0.894 & 0.922 & 0.899  \\
      Slovenian & Slovenian & 0.938 & 0.922 & 0.957 & 0.947  \\
      \hline
      English & Croatian   & 0.824 & 0.724 & 0.822 & 0.725 \\
      English & Slovenian  & 0.830 & 0.719 & 0.848 & 0.736 \\
      Croatian & English   & 0.759 & 0.627 & 0.782 & 0.657 \\
      Croatian & Slovenian & 0.880 & 0.802 & 0.912 & 0.840 \\
      Slovenian & English &  0.741 & 0.578 & 0.794 & 0.648 \\
      Slovenian & Croatian & 0.861 & 0.773 & 0.891 & 0.810 \\
      \hline
\end{tabular}
\caption{The embeddings quality measured on the dependency parsing task. Results are given as UAS and LAS for CroSloEngual BERT and BERT-base-multilingual-cased (mBERT).}
\label{tab:depparse1}
 \end{center}
\end{table}

\begin{table}[!h]
\begin{center}
\begin{tabular}{llrr|rr}
      Train & Test& \multicolumn{2}{c}{mBERT} & \multicolumn{2}{c}{FinEst} \\
      language & language & UAS & LAS & UAS & LAS \\
      \hline
      English & English & 0.917 & 0.894 & 0.918 & 0.895  \\
      Estonian & Estonian & 0.880 & 0.848 & 0.909 & 0.882  \\
      Finnish & Finnish & 0.898 & 0.867 & 0.933 & 0.915  \\
      \hline
      English & Estonian & 0.697 & 0.531 & 0.768 & 0.591 \\
      English & Finnish & 0.706 & 0.561 & 0.781 & 0.624 \\
      Estonian & English & 0.633 & 0.492 & 0.726 & 0.567 \\
      Estonian & Finnish & 0.784 & 0.695 & 0.864 & 0.801 \\
      Finnish & English &  0.543 & 0.433 & 0.684 & 0.558 \\
      Finnish & Estonian & 0.782 & 0.691 & 0.852 & 0.778 \\
      \hline
\end{tabular}
\caption{The embeddings quality measured on the dependency parsing task. Results are given as UAS and LAS for FinEst BERT and BERT-base-multilingual-cased (mBERT).}
\label{tab:depparse2}
 \end{center}
\end{table}



\section{Conclusion}
\label{sec:conclusions}

We built two large pretrained trilingual BERT-based masked language models, Croatian-Slovenian-English and Finnish-Estonian-English. We showed that the new CroSloEngual and FinEst BERTs perform substantially better than massively multilingual mBERT on the NER task in both monolingual and cross-lingual setting. The results on POS tagging and DP tasks show considerable improvement of the proposed models for several monolingual and cross-lingual pairs, while they are never worse than mBERT. 

In future, we plan to investigate different combinations and proportions of less-resourced languages in creation of pretrained BERT-like models, and use the newly trained BERT models on the problems of news media industry.

\section*{Acknowledgments} 
The work was partially supported by the Slovenian Research Agency (ARRS) core research programme P6-0411.
This paper is supported by European Union's Horizon 2020 research and  innovation programme under grant agreement No 825153, project EMBEDDIA (Cross-Lingual Embeddings for Less-Represented Languages in European News Media).
Research was supported with Cloud TPUs from Google's TensorFlow Research Cloud (TFRC).

%
\bibliographystyle{plainnat}
\bibliography{newbert}

\end{document}